\documentclass{article}

\usepackage{amsmath, amssymb}

\usepackage{todonotes}
\usepackage{microtype}
\usepackage{graphicx}
\usepackage{subcaption}
\usepackage{caption}
\usepackage{booktabs} 
\usepackage{multirow}

\usepackage{hyperref}

\usepackage[accepted]{icml2020}

\icmltitlerunning{Continual Learning in Human Activity Recognition Systems}

\begin{document}

\twocolumn[
\icmltitle{Continual Learning in Human Activity Recognition: \\
           an Empirical Analysis of Regularization}




\icmlsetsymbol{equal}{*}

\begin{icmlauthorlist}
\icmlauthor{Saurav Jha}{sta}
\icmlauthor{Martin Schiemer}{sta}
\icmlauthor{Juan Ye}{sta}

\end{icmlauthorlist}

\icmlaffiliation{sta}{School of Computer Science, University of St Andrews, St Andrews, Scotland}

\icmlcorrespondingauthor{Saurav Jha}{sj84@st-andrews.ac.uk}
\icmlcorrespondingauthor{Martin Schiemer}{ms400@st-andrews.ac.uk}

\icmlkeywords{Machine Learning, ICML}

\vskip 0.3in
]



\printAffiliationsAndNotice{} 

\begin{abstract}
Given the growing trend of continual learning techniques for deep neural networks focusing on the domain of computer vision, there is a need to identify which of these generalizes well to other tasks such as human activity recognition (HAR). As recent methods have mostly been composed of loss regularization terms and memory replay, we provide a constituent-wise analysis of some prominent task-incremental learning techniques employing these on HAR datasets. We find that most regularization approaches lack substantial effect and provide an intuition of when they fail. Thus, we make the case that the development of continual learning algorithms should be motivated by rather diverse task domains.

\end{abstract}

\section{Introduction}
The field of continuous learning for neural networks tries to develop algorithms that mimic the mammalian ability to incrementally learn new experiences without deterioration of older ones.
Sensor-based human activity recognition (HAR) aims to autonomously categorize human activities using a  range of sensors such as binary and proximity sensors, accelerometers, etc. to gather information about changes of state or physical activities. The use cases are manifold, ranging from smart homes \cite{Zhang2020} to disease diagnosis \cite{Afonso2019}. HAR's potential benefits from \textit{continuous} learning (often referred to as \textit{lifelong/incremental} learning) are obvious: humans dynamically change their behavior and even develop new activities. Hence, algorithms must adapt to such ever changing diverse behaviors to prevent service quality degradation \cite{Ye2019}. One of the main stepping stones for continuous learning is that learning a new task interferes with previously acquired knowledge - a phenomenon known as \textit{catastrophic forgetting} (CF) \cite{McCloskey1989a}. In general, we would prefer that models are \textit{stable} enough to retain knowledge while being \textit{plastic} enough to incorporate new information \cite{Mermillod2013a}. In this paper, we address techniques that try to alleviate CF through regularization.

Algorithms leveraging regularization attempt to alleviate forgetting through the restriction of updates on network parameters. A substantial amount of these have achieved significant progress on image datasets, it is important to verify their generalization capabilities to other domains such as HAR which is marked by: (i) \textit{dataset imbalance} -- frequencies of activities can vary a lot with some being recurring while others rare; (2) \textit{inter-class similarity} -- activities might resemble each other thus forming overlapping inter-class boundaries; (3) \textit{intra-class diversity} -- an activity can be performed in different ways; and (4) \textit{resource constraints} -- most HAR systems are deployed on memory and computation-constrained devices such as wearables. Characteristics of the sensor datasets used, can be found in Appendices~\ref{ap:similarity} and \ref{ap:classimbdataset}.



The main contribution of our work lies in assessing the applicability and pitfalls of notable continual learning techniques on HAR\footnote{Code will be made available at \url{https://github.com/srvCodes/continual-learning-benchmark}.}. Even though a high volume of continual learning techniques have been proposed in recent years, we focus on regularization and memory replay (MemR) techniques. We are not considering dynamic architecture approaches as HAR systems might not get to see a large number of classes (e.g., around 10-30).

We select five regularization-based methods, which range from classic methods such as LwF~\cite{Li2016} and EWC~\cite{Kirkpatrick2016}, to more recent methods such as  MAS~\cite{Aljundi2018}, LUCIR~\cite{Hou2019} and ILOS~\cite{He2020}. We assess these techniques on two third-party, publicly available datasets that are representative in two common sensor families: \textit{accelerometer} and \textit{ambient sensors}. Through an empirical evaluation on these datasets, we conclude that the regularization terms often have little or even detrimental effect in our scenarios (esp. together with memory replay) and may sometimes be worse than the lower boundary of applying plain cross-entropy (CE) loss. 

\section{Techniques}
This section will briefly introduce the regularization terms whose description can be found in Appendix~\ref{ap:losses}. \textit{LwF} employs knowledge-distillation (KD) loss \cite{Hinton2015} to continuous learning with an objective of maintaining the logits of an incremental step model similar to its predecessor. \textit{EWC} approximates the posterior distribution of network parameters and uses it to identify their importance and penalize their updates. Rotated EWC (RWC) \cite{Liu2018} improves upon EWC by addressing its assumption that fisher information matrix in the network's parameter space are diagonal. Since this is often not the case, they rotate the parameter space in a manner that it does not alter the feed-forward response of the network. \textit{MAS} calculates the importance of parameters by approximating the change in the network output caused by perturbations in parameters due to training on the new task data. \textit{LUCIR} introduces two loss terms: \textit{less forget constraint} (DIS) and \textit{margin ranking} (MR)\footnote{MR only applies to memory replay since in-memory samples are used to distance class embeddings.} with the goals of preventing rotation of old class embeddings and reducing ambiguities between old and new classes. \textit{ILOS} modifies the CE loss by replacing the new model's logits for old classes with those adjusted proportionately between new and previous model. They coin the resultant loss as cross-distillation loss. We also consider lower bound as the model trained with CE loss and upper bound as the offline training with all tasks at the same time. The CE loss is defined as $\mathcal{L}_{CE}(y, \hat{y}) = -\sum y \log \hat{y}$, where $y$ and and $\hat{y}$ are the ground truth and output logits for an input sample.

\section{Experimental Setup}\label{sec:exp}
Our main objective is to assess which type of regularization term is effective for continual learning on sensor-based HAR and to what degree.  

\subsection{Datasets}
We select two datasets from the sensor-based HAR community. The first dataset \textbf{(WS)} was collected on 32 ambient passive infra-red sensors by a smart home testbed at the Washington State University's CASAS.\footnote{\hyperlink{http://ailab.wsu.edu/casas/datasets/}{http://ailab.wsu.edu/casas/datasets/}} It includes 9 imbalanced 
activities: cooking, eating, leaving/entering the house, living room activity, toilet use, mirror, reading, sleeping, and working. The second dataset is \textbf{DSADS} -- Daily and Sports Activities Dataset~\cite{Altun.2010}. It is a balanced dataset with 19 activities that include sitting, running on a treadmill, exercising on a stepper, and rowing among others - each of which is performed by 8 subjects for 5 minutes with 5 accelerometer units on a subject's torso, right arm, left arm, right leg and left leg. Circumventing the topic of feature extraction, we work on the features already extracted by prior work instead of the raw spatio-temporal sensor data. For DSADS, we use a  version processed by \citet{Wang2018} which extracts 27 features (including mean, standard deviation, and correlations on axes) on each sensor. For WS, we use those generated by \citet{Fang2020}.

\subsection{Evaluation Process}
Considering task order-sensitivity in continual learning paradigms \cite{2019arXiv190209432Y}, we evaluate the techniques on 30 task sequences while updating the parameters on every incoming task. Each task is coupled with two randomly sampled classes, thus contributing to a sequence length of $|C|/2$, where $|C|$ is the total number of classes in the dataset.

We perform a stratified train-test split of 70/30 on WS dataset while for DSADS, we split data on participants; \textit{i.e.,} we use data from 70\% of subjects for training and the remaining 30\% for testing. After training, we retain $\frac{S*|C|}{|C|_{seen}}$ random samples per class in the memory to be replayed at further incremental training steps. $S$ is determined by the memory constraint of the HAR system and $|C|_{seen}$ is the number of classes observed till the current incremental step. 

\begin{table*}[t]
    \centering
    \caption{Performance comparison of different regularization terms. \footnotesize{\textit{wo.} and \textit{w.} refer to \textit{without} and \textit{with} memory replay respectively.}}
    \label{tab:regcompare}
    \includegraphics[width=0.8\textwidth]{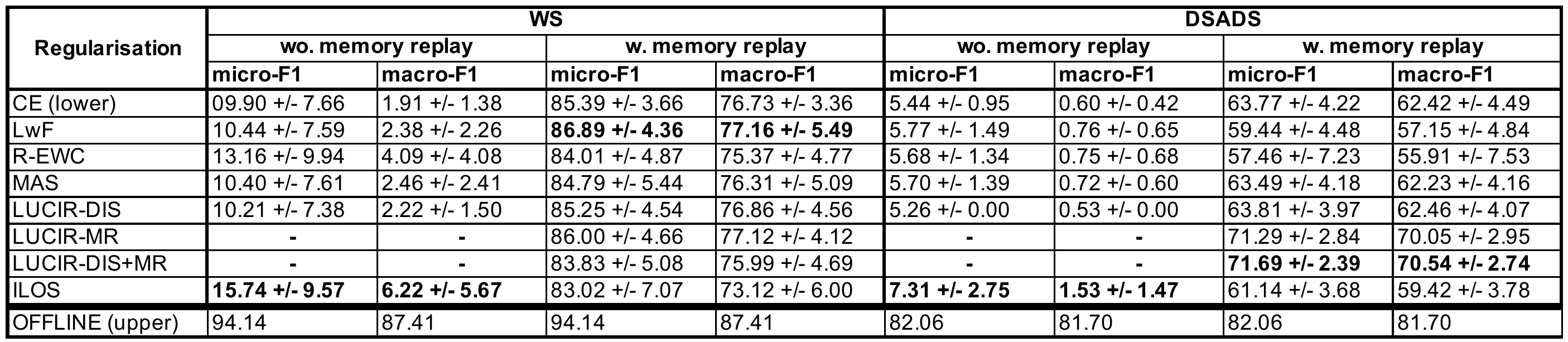}
\end{table*}

\begin{figure*}[t]
    \centering
    \includegraphics[width=0.98\textwidth]{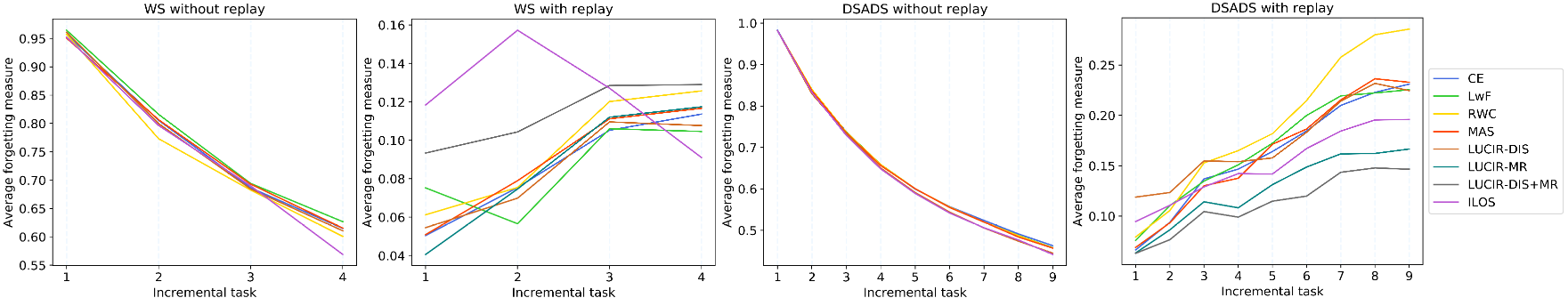}
    \caption{Forgetting measure comparison of losses with and without memory replay.}
    \label{fig:forgetting}
\end{figure*}

\begin{figure*}[t]
    \centering
    \includegraphics[width=0.98\textwidth]{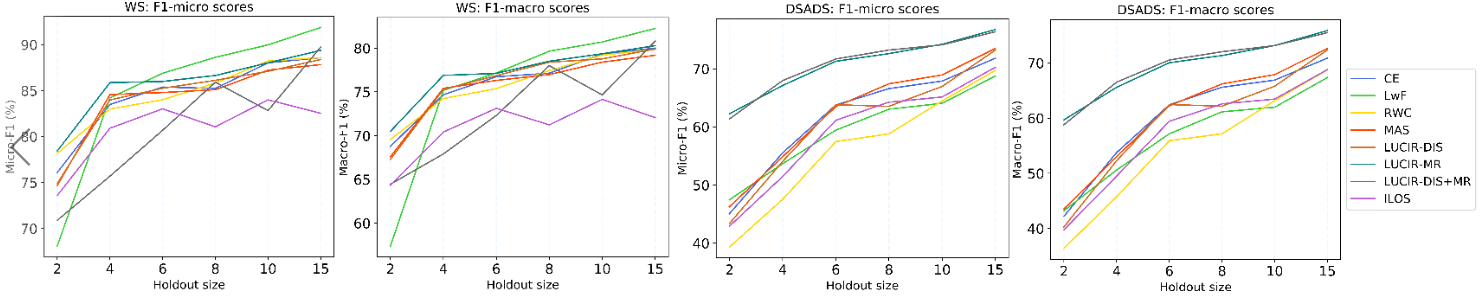}
    \caption{F1 performance comparison of different in-memory sizes. WS F1 scores on the left, DSADS on the right.}
    \label{fig:holdout}
\end{figure*}

\subsection{Evaluation Metrics}
Upon arrival of a new task $k$, we compute four types of accuracy: \textit{base} and \textit{old} class accuracies measure performance on the very first ($0^{th}$) task and the tasks $\{1,..,k-1\}$ henceforth, thus indicating the \textit{stability} of the model;  \textit{new} accuracy measures the performance on the current task thus indicating the \textit{plasticity} of the model; and \textit{overall} accuracy considers all the tasks learned so far, and implies the \textit{stability-plasticity} balance of the model. The accuracy is measured in micro-F1 scores. Given the imbalanced class distribution in a real-world HAR scenario (see Appendix \ref{ap:classimbdataset}), we additionally report macro-F1 scores.

For discerning the preservation of existing knowledge, we calculate the \textit{forgetting measure} proposed by \citet{Chaudhry2018} which for task $k$ is the difference between its maximum accuracy seen so far and the current accuracy averaged over $\{1,..,k-1\}$ tasks: $F_k=\frac{1}{k-1}\sum_{j=1}^{k-1}a_{k,j,max}-a_{k,j}$.

\subsection{Model Configuration and Hyperparameter Tuning}
Irrespective of their original works, we maintain a common network architecture across all our experiments as a fair comparison premise. We use fully-connected feed-forward networks with the following specifications optimized through extensive grid search: (1) DSADS: 3 hidden layers of sizes [202, 202, 101], and (2) WS: 2 hidden layers of sizes [32, 16, 16]. Each network has a single output head that gets extended on each incoming task to accommodate for new classes.

We perform a further search for technique-specific hyperparameters, detailed in Appendix~\ref{ap:hyper}.  It is worth noting that our LUCIR-based losses employ L2 normalization of the output logits of FC layer rather than the cosine normalization which offers a significant boost to performance in the original work of \citet{Hou2019}. This compliments our fair premise assumption of assessing regularization alone. 

\section{Results}
\paragraph{Fixed holdout size:} Table~\ref{tab:regcompare} compares the micro and macro-F1 scores of regularization terms on WS and DSADS with and without MemR. The replay-based scores use $S=6$ which we assume to be small enough to be held in a resource-constrained device and large enough to deliver decent performance. We find that most of the regularization techniques when devoid of replay only achieve the naive accuracy of baseline CE. In this scenario, ILOS with a direct influence of logits from the old model performs better than the rest where the models learn to  align them as training progresses. When aided with replay, we find that CE alone beats most of the other techniques on both the datasets. For example, the improvements over the baseline CE approach remain within 1\% on the WS dataset and within 8\% on the DSADS dataset on both micro and macro-F1. We further observe that LUCIR's less-forget-constraint (DIS) does not provide a strong effect.

In terms of \textit{task order-sensitivity} without MemR, LUCIR-DIS  with the least average standard deviation has a clear win over the rest of the methods while ILOS and RWC offer less robustness. When MemR is used, the picture is more diverse between datasets as LwF and LUCIR-DIS+MR are the most stable for WS and DADS respectively.

We  see that the differences between the F1 micro and macro scores vary across the methods. For the results without MemR, the micro scores are multiple times higher with CE being the most divergent (518\% and 907\%) while ILOS the least (253\% and 448\%). Table~\ref{tab:F1Comp} in Appendix~\ref{ap:diffF1} presents the divergence scores between F1-micro and macro. From this, we conclude that the regularization methods help in learning fairer distributions of classes. In contrast, the advantage of the regularization methods with MemR is less apparent with no big difference to CE.
\paragraph{Forgetting:} Figure \ref{fig:forgetting} depicts the stark contrast of forgetting scores ($F$) of replay-assisted techniques to those without replay. Without MemR, $F$ decreases sharply below 1.0 across all methods as the  learning progresses beyond task 1. Although the strikingly high forgetting scores stipulate catastrophic forgetting on earlier tasks, these further shed light into the \textit{stability} of techniques that are devoid of replay as their forgetting diminish with the arrival of further diverse tasks. On the other hand, the contribution of the regularization terms improve from being null to modest with MemR. In particular, we observe a threshold number of incremental tasks for replay-assisted methods following which the inertia of forgetting dampens. ILOS, LUCIR-MR and LUCIR-DIS+MR attain this threshold much earlier than CE. The high forgetting scores of RWC conform to the finding of \citet{kemker2018measuring} stating EWC-based methods are poor at learning new categories incrementally. LwF and ILOS start with relatively larger forgetting scores whose slope alleviates with further incremental steps. In contrast, margin ranking-based techniques have lesser overall forgetting scores, which accords with greater inter-class separation between and old and new classes.

\begin{figure}[ht]
    \centering
    \begin{subfigure}{\columnwidth}
        \includegraphics[width=\columnwidth]{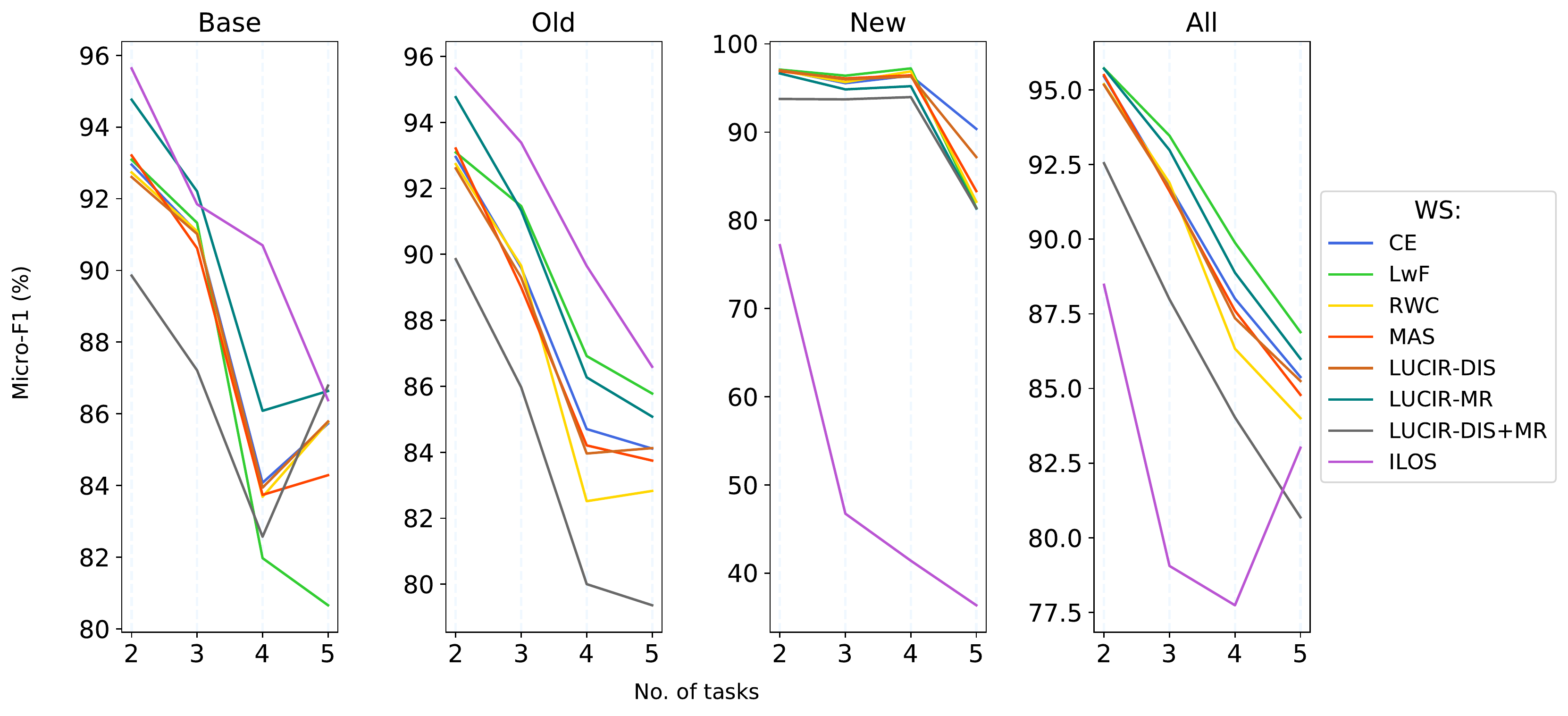}

    \end{subfigure}%
            \vspace{-1\baselineskip}
    \begin{subfigure}{\columnwidth}
        \includegraphics[width=\columnwidth]{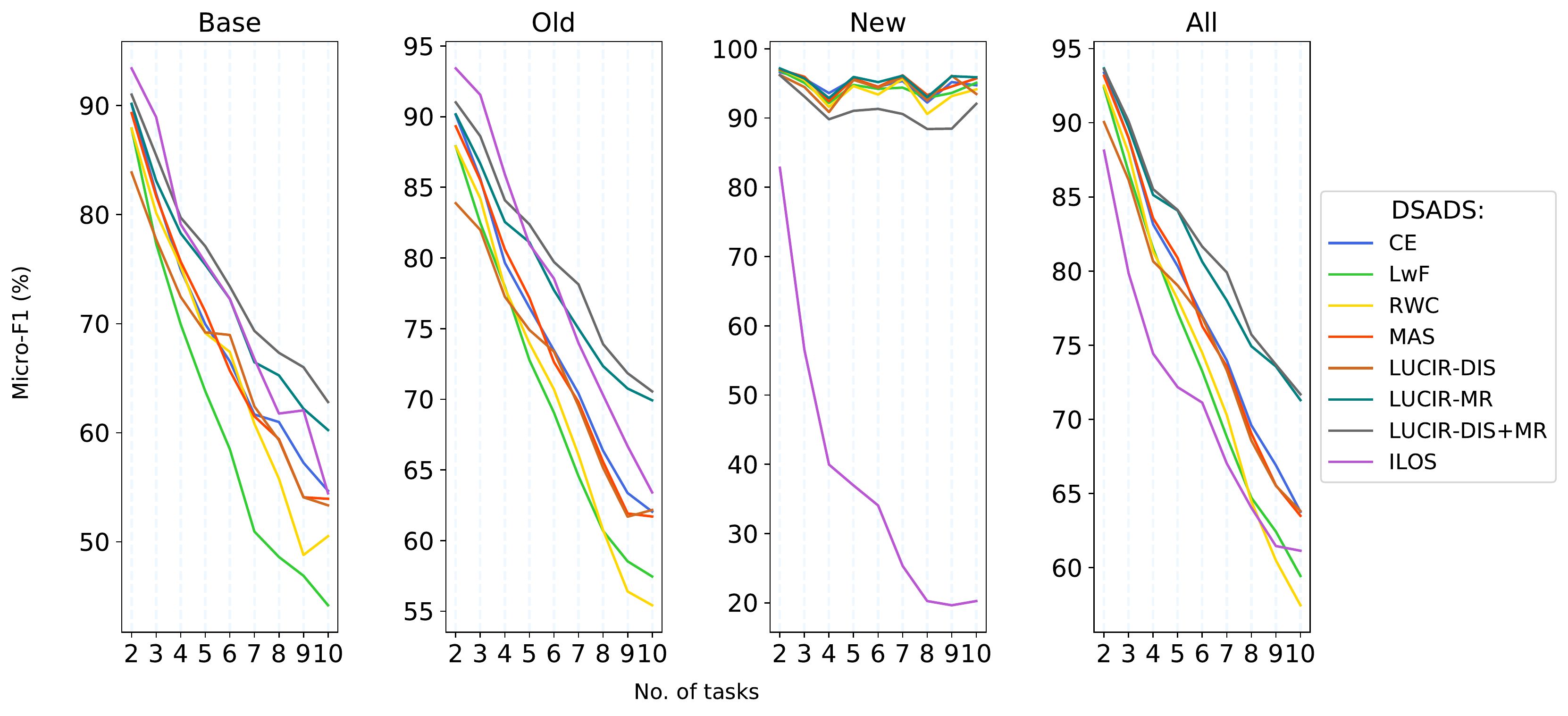}

        \end{subfigure}

\caption{Accuracies detailed by base, old, new and all classes per incremental task.}    \label{fig:deepacc}
\end{figure}

\paragraph{Performance across base, new and old classes:} While LUCIR-DIS+MR and LwF outperform other techniques on DSADS and WS respectively on overall tasks, an observation of the base, new and old class accuracies in Figure \ref{fig:deepacc} offers additional insights into how different techniques respond to the \textit{plasticity}-\textit{stability} trade-off. ILOS, for instance, consistently performs poorer on new classes across both the data sets. However, the maximum scores of ILOS on base and old classes make it more robust to interferences due to new knowledge hence showing that even a direct tuning of the new model's logits based on the previous model can help surpass complex regularization operations. Together with this and the divergences between F1 macro and micro scores, we assume that ILOS's restrictiveness for new classes actually harms the learning when used in conjunction with MemR. We also observe that margin-ranking based methods (LUCIR-MR and LUCIR-DIS+MR) perform poorer than others on new tasks but are robust at preserving old knowledge. An intuitive explanation to this could be the design of MR that reinforces the model's confidence at recognizing ground truth embeddings for old class samples following multiple incremental training steps.
\paragraph{Varying holdout sizes:} Drawing inspiration from the superior performance of replay-assisted learning, we further evaluate these for $S \in \{2,4,6,8,10,15\}$. Figure \ref{fig:holdout} shows that even a small number of replay samples can yield a huge leap in the performance of the techniques than when ran without replay (Table \ref{tab:regcompare}). For WS, we see that the majority of the methods react in the same way to memory upgrades as their F1 scores start with a similar slope and reach a comparable pace starting from step 4. A few of them stick out: ILOS which  mostly has a lower score than the rest but the slope is similar. LUCIR-DIS+MR reaches the level of most methods later and is less stable. We attribute these anomalies to the imbalanced sample distribution of WS. Looking at the results on DSADS which has perfectly balanced class distribution, this seems to be the case as we can see an almost linear increase in score with rising memory. Thus, we conclude that the working of regularization terms is more dependent on dataset characteristics than on the available memory.

\section{Conclusion}
In this paper, we have shown that the well known continuous learning regularization terms have no or only  limited effect in human activity recognition scenarios when used with or without memory replay.

Memory replay, in particular, overshadows the value of regularization and some techniques even adversely affect the learning process. Most importantly, we advocate that the direction of continual learning research should not only focus vision tasks but also target other domains with diverse data distribution and resource constraints. 


\newpage
\bibliography{main}
\bibliographystyle{icml2020}

\newpage
\appendix
\section{Loss Function Terms}\label{ap:losses}

\paragraph{LwF} uses knowledge distillation loss to approximate the output of the original network:
\begin{equation}\label{eq:kd_loss}
     \mathcal{L}_{KD}(y_o, \hat{y}_o) = -\sum\limits_{i=1}^{l}y_o^{'(i)}\log\hat{y}_o^{'(i)}, 
\end{equation}
where $l$ is the number of class labels, and $y_o^{'(i)}$ and $\hat{y}_o^{'(i)}$ are temperature-scaled \textit{recorded} and \textit{current} probabilities of the sample on a label $l$. The loss $\mathcal{L}_{KD}$ is combined with the cross-entropy loss on new task samples to form the \textit{cross-distillation} loss: 
\begin{equation}
\label{crossdis}
    \mathcal{L}(y_n,\hat{y}_n,y_o, \hat{y}_o) = \lambda_{}o\mathcal{L}_{KD}(y_o, \hat{y}_o) + \mathcal{L}_{CE}(y_n, \hat{y}_n)
\end{equation}
where $\lambda_o$ is a loss balance weight computed as the ratio of old classes to total observed classes. A larger $\lambda_o$  favors the old task performances over new task.
\paragraph{EWC} assumes that if a dataset $D$ consists of two independent tasks $A$ and $B$, then the importance of parameters of the model is modeled as the posterior distribution $log p(\theta|D) = log p(D_B|\theta) + log p(\theta|D_A) - log p(D_B)$. $p(\theta|D_A)$  suggests which parameters are important to task $A$. The true posterior probability $p(\theta|D_A)$ is intractable, and thus it is estimated via Laplace approximation~\cite{MacKay1992} with precision determined by the Fisher Information Matrix (FIM). The loss function for EWC is defined as: 
\begin{equation}\label{eq:ewc}
   \mathcal{L}(\theta) = \mathcal{L}_B(\theta)+\sum\limits_i\frac{\lambda}{2}F_i(\theta_i-\theta^*_{A,i})^2, 
\end{equation}
where $\mathcal{L}_B$ is the loss on task B, $\lambda$ indicates the importance of the old task with respect to the new task, and $i$ is the parameter index. \textbf{RWC} improves upon EWC by reparameterizing $\theta$ through rotation in a way that it does not change outputs of the forward pass but the FIM computed from gradients during the backward pass is approximately diagonal. 

\paragraph{MAS} considers approximating the importance of a network's parameters by learning the sensitivity of the objective function to a parameter change; \textit{i.e.,} given a data point $x_k$ whose network output is $F(x_k; \theta)$, a change in the network output caused by a small perturbation $\delta = \{\delta_{ij}\}$ in the parameters $\theta=\{\theta_{ij}\}$ can be approximated as: $F(x_k;\theta+\delta) - F(x_k;\theta) \approx \sum\limits_{i,j}g_{ij}(x_k)\delta_{ij}$, where $g$ is the gradient with respect to the parameter $\theta$, and $g_{ij}(x_k) = \frac{\partial(F(x_k,\theta)}{\partial\theta_{ij}}$. Accumulating gradients over all the data points, the importance weight on a parameter $\theta_{ij}$ can be computed as: $\Omega_{ij} = \frac{1}{N}\sum\limits_{k=1}^N||g_{ij}(x_k)||$.
While learning a new task, MAS then defines the loss function as:
\begin{equation}\label{eq:mas}
    \mathcal{L}(\theta) = \mathcal{L}_n(\theta) + \frac{\lambda}{2}\sum\limits_{i,j}\Omega_{ij}(\theta_{ij}-\theta^*_{ij})^2, 
\end{equation}
where $\mathcal{L}_n(\theta)$ is the loss on the new task, $\theta_{ij}$ and $\theta^*_{ij}$ are the new and old network parameters, and $\lambda$ is a hyperparameter that varies with the dataset.

\paragraph{LUCIR} primarily targets the \textit{class imbalance}  arising due to a small amount of in-memory samples of old tasks and a large amount of samples of new tasks in the data of an incremental training step. This is tackled using two kinds of losses: (i) \textit{less-forget constraint} loss ($\mathcal{L}_{dis}^G$)  is introduced to preserve the spatial configuration of old classes' embeddings by encouraging the features extracted from the new model to be rotated in the direction similar to those of the old model, \textit{i.e.,} 
$\mathcal{L}_{dis}^G(x)= 1 - \langle\Tilde{f}^*(x),\Tilde{f}(x)\rangle$, where $\Tilde{f}(x)$ and $\Tilde{f}^*(x)$ are normalised features extracted by the new and the old model respectively, and $\langle v_1,v_2 \rangle$ denotes the cosine similarity between the vectors $v_1$ and $v_2$; (ii)  \textit{margin ranking} ($\mathcal{L}_{mr}$) loss is used to enhance inter-class separation by pushing the ground-truth old classes for each in-memory sample $x$ far from all new classes it is confused with. To achieve this, the logits of ground-truth classes of $x$ are treated \textit{positive} while the logits of top-K classes that $x$ is most confused with are treated as hard \textit{negatives}, \textit{i.e.,}   $\mathcal{L}_{mr}(x)=\sum\limits_{k=1}^K\max(m-\langle\Tilde{\theta}(x),\Tilde{f}(x)\rangle+\langle\Tilde{\theta}^k.\Tilde{f}(x)\rangle,0)$. The loss function resulting from the combination of $\mathcal{L}_{dis}^G$ and $\mathcal{L}_{mr}$ can be given as:
\begin{equation}\label{eq:LUCIRcomb}
    L=\frac{1}{|N|}\sum\limits_{x\in N}(L_{ce}(x)+\lambda L_{dis}^G(x))+\frac{1}{|N_o|}\sum\limits_{x\in N_o}L_{mr}(x), 
\end{equation}
where $N$ is a training batch drawn from $X$ and $N_o$ represents the reserved old samples. $\lambda$ is a hyperparameter that says how much knowledge of the previous model needs to be preserved depending on how many new classes are introduced and is computed by multiplying a fixed $\lambda_{base}$ with the squared root of the fraction of new and old classes; \textit{i.e.,} $\lambda=\lambda_{base}\sqrt{|C_N|/|C_o|}$.

\paragraph{ILOS} uses an accommodation ratio $0 \leq \beta \leq 1$ to adjust the proportion of logits from the current model and the previous model:
\begin{equation}\label{eq:ilos}
    \Tilde{o_k} =
    \begin{cases}
   \beta o_k + (1 - \beta)\hat{o_k} &  1 \leq k \leq n \\
    o_k,    & n+1 \leq k \leq n+m
\end{cases}
\end{equation}

where $n$ is the number of classes observed till previous task, $m$ is the number of classes in the current task, $\Tilde{o_k}$ are the adjusted output logits and $\hat{o_k}$ are the output logits from the FC layer of the previous model. The adjusted norms of old classes are thus inclined either towards the range of norms of old classes of the current model or that of the previous model. The degree of this inclination is proportional to the magnitude of $\beta$. While $\mathcal{L}_{KD}$ in Equation \ref{crossdis} is still calculated using $o_k$, $\mathcal{L}_{CE}$ is now based on $\Tilde{o_k}$ instead of $o_k$. 

\section{Inter-class Similarity}\label{ap:similarity}
\begin{figure}[H]
    \centering
    \includegraphics[width=0.48\textwidth]{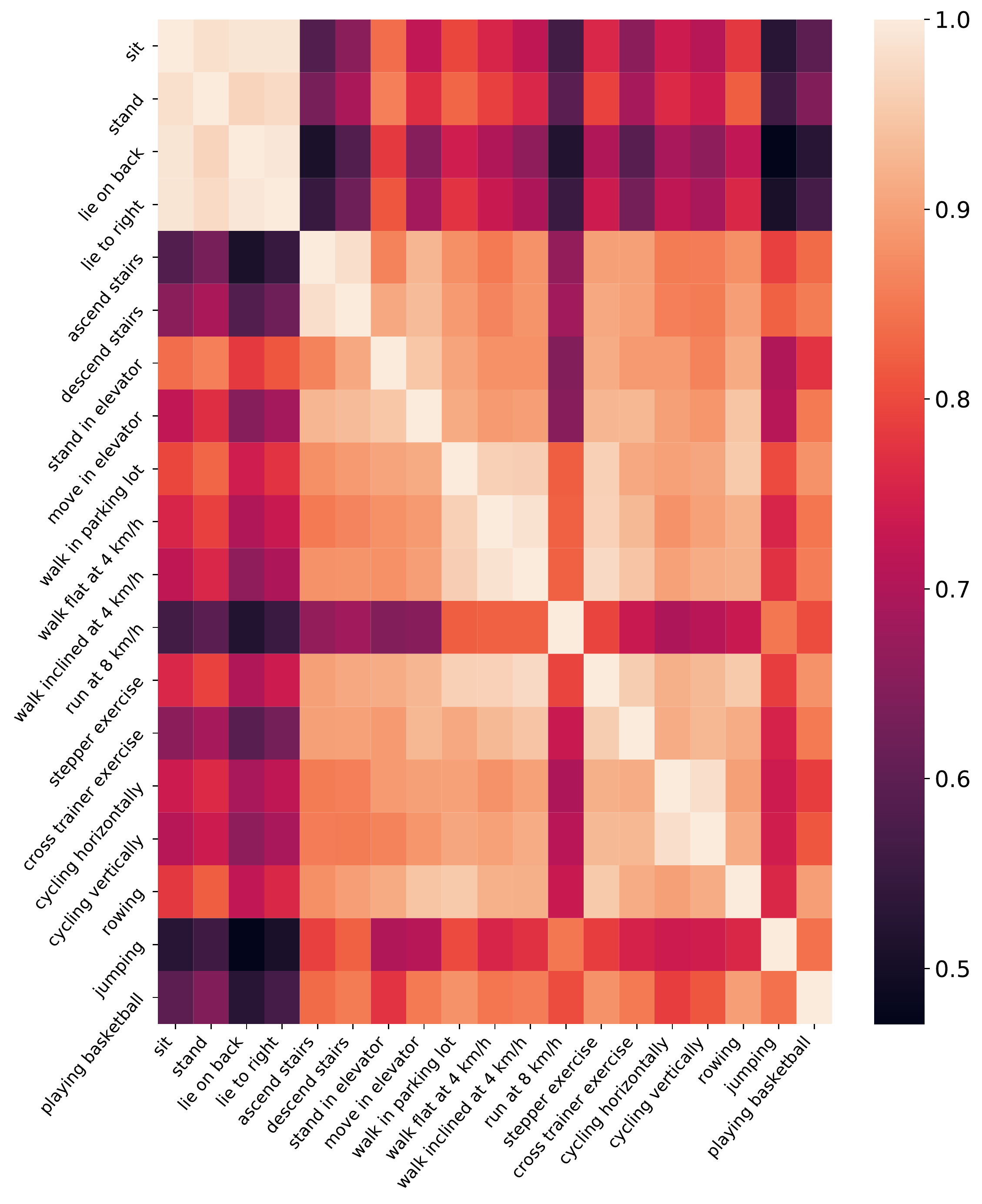}
    \caption{Correlation heatmap of activities in the DSADS accelerometer dataset.}
\label{fig:correlation}\end{figure}

Figure \ref{fig:correlation} shows the correlation among raw features of activities in DSADS. Due to bodily restrictions and subject-specific fashion, different activities might have resemblance in distribution.

\section{Class distribution}
\label{ap:classimbdataset}
\begin{figure}[h!]
    \centering
    \begin{subfigure}{\columnwidth}
        \includegraphics[width=\columnwidth]{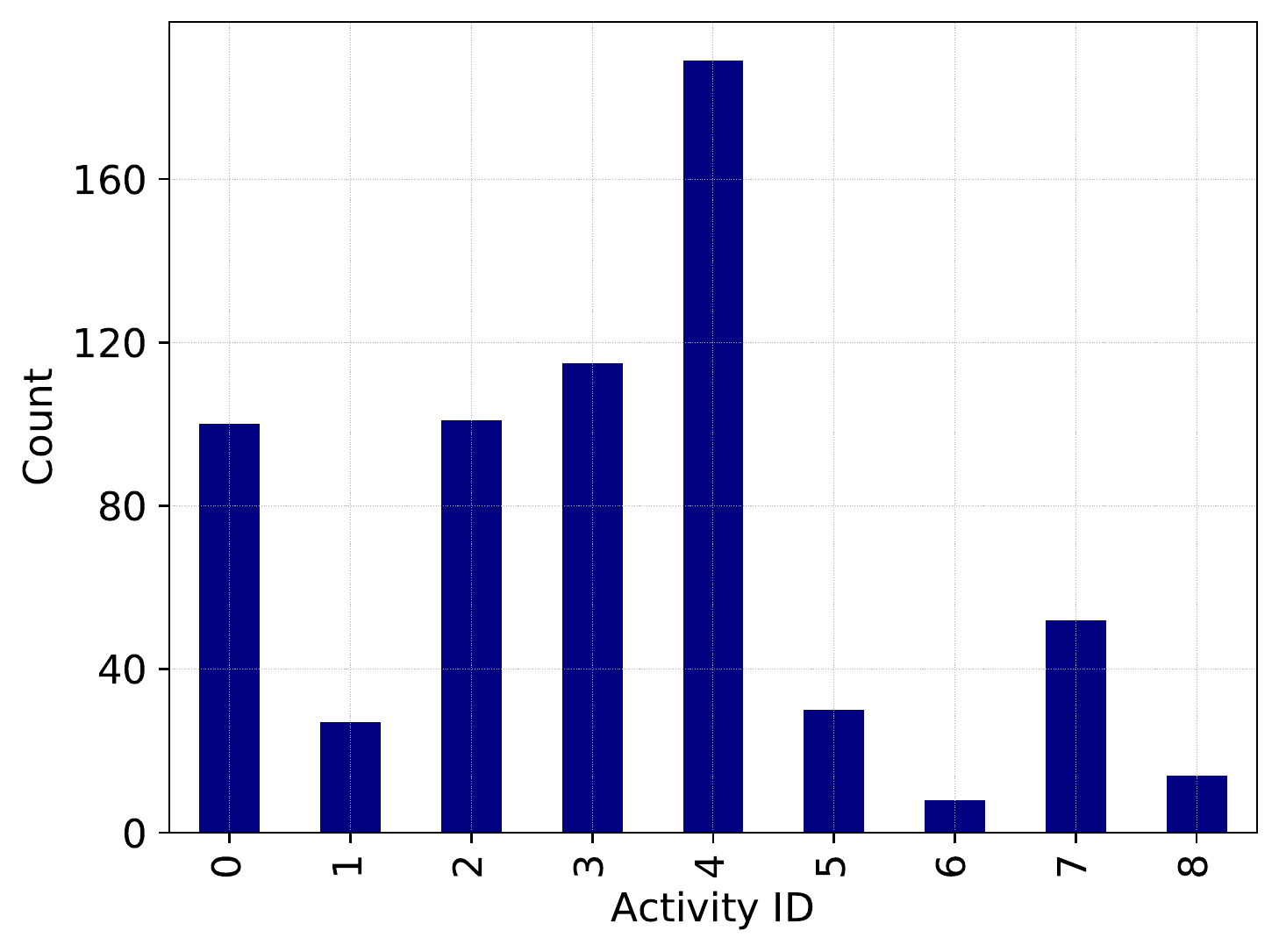}
        \caption{WS}
        \label{fig:gull}
    \end{subfigure}%
    \newline
    \begin{subfigure}{\columnwidth}
        \includegraphics[width=\columnwidth]{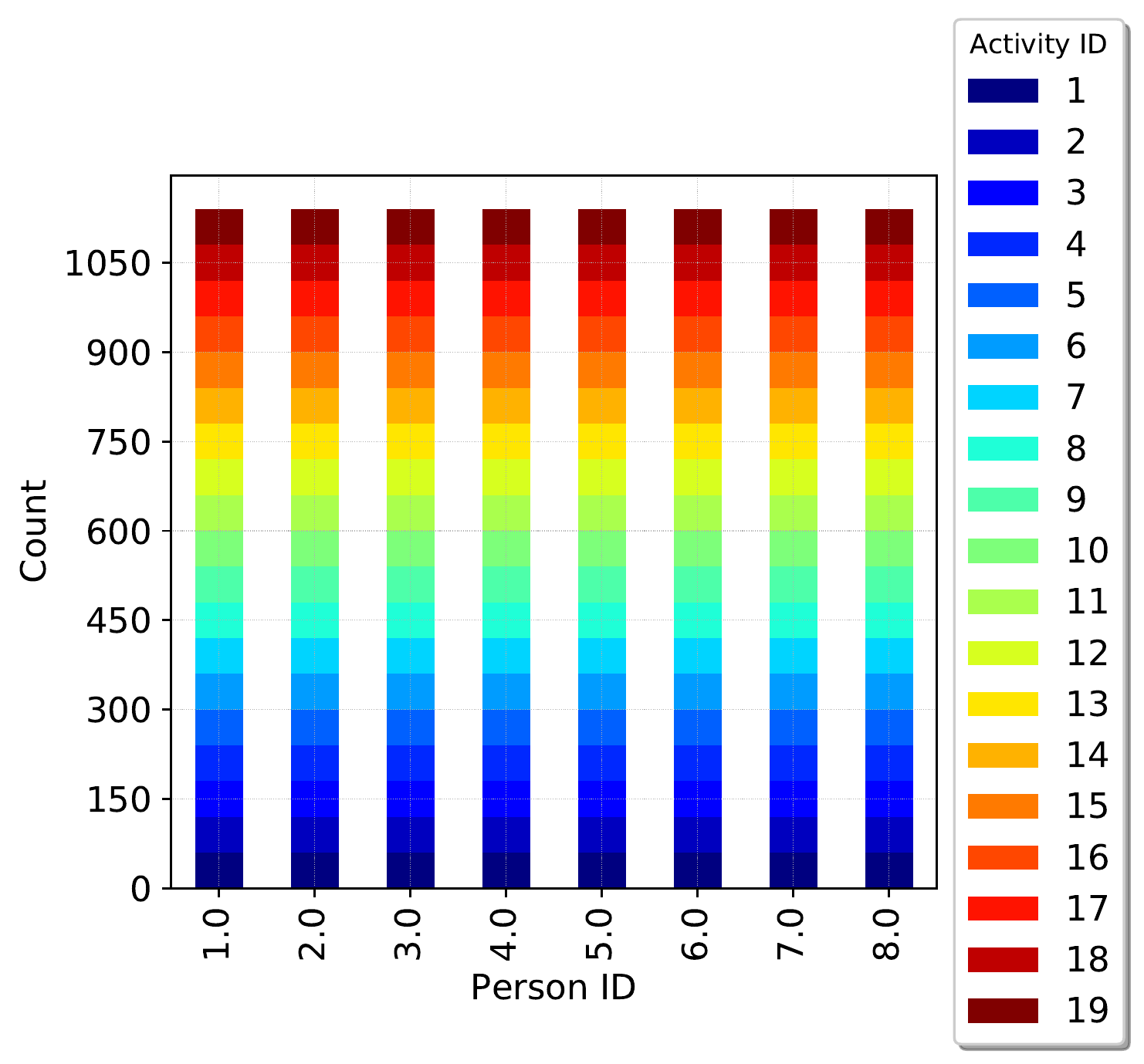}
        \caption{DSADS}
        \label{fig:gull}
        \end{subfigure}
\caption{Frequency distribution of activities in WS and DSADS datasets.}    \label{fig:imablance_label}
\end{figure}

As shown in Figure \ref{fig:imablance_label}, the two datasets used in our work represent two different scenarios of class distribution apart from having captured using different sensor technologies.



\section{Technique-specific hyperparameters} \label{ap:hyper}
This is a brief overview over the hyperparameters derived by grid search. $\lambda$ for LwF, RWC and MAS are set to 1.6, 3 and 0.25 each. LUCIR-based losses use a $\lambda_{base} = 5$, and $m$ and $k$ for LUCIR-MR are set to 0.5 and 2 each.

\begin{table}[H] 
            \caption[-]{Additional experiment hyperparameters}
        	\label{tab:hyper}
        	\centering
        	{\footnotesize
        	\resizebox{\columnwidth}{!}{
        	\begin{tabular}{l|ll} 
        		\toprule
        		Dataset & WS & DSADS\\
        		\midrule
        		Batch size   & 15 & 20\\ 
        		Initial Learning Rate & 0.01 & 0.01\\ 
        		Epochs till Convergence & 200 & 200\\ 
        		Learning Rate Scheduler Step Size (effective after)   & 40 (50)\footnotemark & 50 (50)\\ 
        		Weight Decay Rate   & 1.00E-04 & 1.00E-04
        	\end{tabular}
        	}
        	}
        	
            \end{table}
\footnotetext{For example, learning rate for training on WS reduces by a factor of 0.01 after 90, 130 and 170 epochs.}

\section{Relative Divergence between F1 Micro and Macro Scores}\label{ap:diffF1}

\begin{table}[H] 
            \caption[-]{F1-Micro / F1-Macro in percent (\%).}
        	\label{tab:F1Comp}
        	\centering
        	{\footnotesize
        	\resizebox{\columnwidth}{!}{
        	\begin{tabular}{l|ll|ll} 
        		\toprule
        		Method & WS Blank & WS Memory Replay & DSADS Blank & DSADS Memory Replay\\
        		\midrule
        		CE   & 518.32 & 111.29& 906.67& 102.16\\ 
        		LwF & 438.66& 112.61& 759.21& 104.01\\ 
        		RWC & 321.76& 111.46 & 757.33 & 102.77\\ 
        		MAS & 422.76& 111.11 & 791.67 & 102.02 \\ 
        		LUCIR-DIS   & 459.91 & 110.92 & 992.45 & 102.16 \\
        		LUCIR-MR   & - & 111.51 & - & 101.77 \\
        		LUCIR-DIS+MR & - & 110.32 & - & 101.63\\
        		ILOS & 253.05 & 113.54 & 477.78 & 102.89
        	\end{tabular}
        	}}
        	
            \end{table}

\end{document}